\documentclass[sigconf]{acmart}

\setcopyright{none}
\renewcommand\footnotetextcopyrightpermission[1]{}
\acmYear{2027}
\acmConference[KDD '27]{KDD 2027 AI for Sciences Track}{August 2027}{San Jose, CA, USA}
\acmDOI{}
\acmISBN{}

\usepackage{amsmath}
\usepackage{booktabs}
\usepackage{graphicx}
\usepackage{subcaption}
\usepackage{algorithm}
\usepackage{algpseudocode}
\usepackage{xcolor}
\usepackage{multirow}
\usepackage{placeins}
\usepackage{tikz}
\usetikzlibrary{arrows.meta,positioning,fit,calc}
\definecolor{pcblue}{HTML}{0072B2}
\definecolor{pcorange}{HTML}{E69F00}
\definecolor{pcgreen}{HTML}{009E73}
\definecolor{pcpurple}{HTML}{CC79A7}
\definecolor{pcgray}{HTML}{666666}
\definecolor{pcdark}{HTML}{222222}

\title{PsyWorld: Profile-Conditioned World Modeling for Human Behavior Prediction}

\author{Ziyuan Qin}
\email{ziyuan.qin@emory.edu}
\affiliation{%
  \department{Department of Computer Science}
  \institution{Emory University}
  \city{Atlanta}
  \state{Georgia}
  \country{USA}}

\author{Yang Qiao}
\email{yang.qiao@emory.edu}
\affiliation{%
  \department{Department of Computer Science}
  \institution{Emory University}
  \city{Atlanta}
  \state{Georgia}
  \country{USA}}

\author{Summer B. Frandsen}
\email{summer.frandsen@emory.edu}
\affiliation{%
  \department{Department of Psychiatry and Behavioral Sciences}
  \institution{Emory University}
  \city{Atlanta}
  \state{Georgia}
  \country{USA}}

\author{Michael T. Treadway}
\email{mtreadway@emory.edu}
\affiliation{%
  \department{Department of Psychiatry and Behavioral Sciences}
  \institution{Emory University}
  \city{Atlanta}
  \state{Georgia}
  \country{USA}}

\author{Liang Zhao}
\email{liang.zhao@emory.edu}
\affiliation{%
  \department{Department of Computer Science}
  \institution{Emory University}
  \city{Atlanta}
  \state{Georgia}
  \country{USA}}

\begin{document}

\begin{abstract}
Understanding how stable psychometric traits and time-varying states jointly
inform behavior is an important problem in computational behavioral science.
Existing personalized agents commonly condition actions directly on a static
profile, leaving realized state transitions unused as an explicit source of
supervision. We introduce a profile-conditioned state world model that uses observed
environment--behavior state transitions to learn transition-informed
representations for behavioral prediction. During training, a posterior encoder
quantizes realized transitions, while a prior predicts the resulting codes from
the psychometric profile and pre-transition context. The prior code and
predicted next state then condition macro-level behavior and simulator-grounded
execution. Thus, the world model provides structured transition supervision
instead of treating personality as a direct action descriptor. We evaluate the
framework in VEET, a controlled behavioral simulator with human trajectories,
and in AgentSociety as a structurally different synthetic environment. A
participant-disjoint split of 240 training, 25 validation, and 48 test
participants evaluates generalization to unseen people. On VEET, the model
reaches 0.904 Top-1 accuracy (95\% CI: [0.895, 0.913]) and 0.0033 Macro KL.
Ablations associate continuous next-state modeling with most of the Top-1 gain,
while discrete codes and prior alignment mainly improve Top-3 and distributional
alignment. These results establish transition-informed world-model supervision
as useful for subject-generalized behavior prediction, without implying causal
psychological mechanisms or calibrated long-horizon state dynamics.
\end{abstract}



\maketitle

\section{Introduction}

Human behavior reflects both stable individual differences and states that vary
within a task. Psychometric profiles describe between-person variation in
stress, motivation, and related traits, whereas recorded task, environmental,
and behavioral variables describe the context in which those tendencies are
expressed. Dense longitudinal evidence shows that trait- and state-level
motivation can exert distinct and interacting effects on effort-based
decisions~\citep{pshyco_background}. For computational behavioral simulation,
this distinction suggests that realized state transitions can provide
supervision beyond a static profile-to-action mapping.

Existing personality-conditioned agents only partially use this structure.
LLM-based social simulators combine persona descriptions with needs, memory, and
planning~\citep{piao2025agentsociety,d2a}, while consistency-oriented training
reduces persona drift across interactions~\citep{abdulhai2025consistentlysimulatinghumanpersonas}.
These systems can produce locally plausible behavior, but profiles are commonly
injected as static prompts, policy inputs, or reward descriptors. A direct
profile-to-action model can therefore capture participant-level associations
without representing within-person changes in task, environmental, and
behavioral state.

The central challenge is to exploit realized state changes during training
without using future information at inference. Realized environment--behavior
state changes provide transition-level supervision associated with subsequent
behavior, but the future state is unavailable before an agent acts. Under
participant-disjoint evaluation, the model must also transfer this structure to
profiles and trajectories never observed during training, rather than memorize
participant-specific action frequencies. We therefore ask whether explicit
state-transition supervision can improve behavioral prediction for previously
unseen participants.

World models provide a natural inductive bias for this question because they
learn state evolution rather than only action association. Latent world models
have enabled agents to learn compact transition structure and reason over
predicted futures~\citep{dreamerv1,dreamerv2,dreamerv3}. Standard world models,
however, are designed primarily for task control and do not explicitly separate
stable psychometric profiles from the time-varying environment--behavior state.
Our formulation makes that distinction explicit and uses next-state prediction
as structured supervision for downstream behavior.

We introduce a \emph{profile-conditioned state world model}. For each
transition, the model separates the stable psychometric profile from the current
environment--behavior state and macro behavior. A training-time posterior
observes the realized state change and assigns it to a vector-quantized
transition code. A rollout-time prior predicts that code using only the profile
and pre-transition context, making the transition-informed representation
available without future leakage. The code and predicted next state condition
macro-level behavior, which is realized through a simulator-specific micro
controller. The world model is therefore central to the
profile--state--behavior factorization, while the evaluated outcome is behavior
prediction rather than independent calibration of long-horizon state dynamics.

We evaluate the formulation in two complementary environments. VEET provides
questionnaire profiles and behavioral trajectories from human participants for
the primary unseen-participant evaluation. AgentSociety provides a structurally
different synthetic LLM-based environment for testing whether the same
factorization can be operationalized across state spaces and action interfaces.
We use a strictly participant-disjoint split of 240 training, 25 validation, and
48 test participants, so every reported VEET test trajectory comes from a
previously unseen person. Ablations separate the effects of continuous
next-state prediction, vector quantization, and rollout-time prior alignment; a
code--trait analysis bounds the interpretation of the learned representation.

Our contributions are:
\begin{itemize}
    \item We formulate subject-generalized human behavior prediction by
    separating stable psychometric profiles from time-varying
    environment--behavior states and using realized transitions as structured
    supervision.
    \item We develop a profile-conditioned state world model whose posterior
    extracts vector-quantized transition codes from observed state changes and
    whose prior predicts them from pre-transition context for leakage-free
    rollout.
    \item On a participant-disjoint VEET split, we show the predictive utility
    of transition supervision and characterize the distinct roles of continuous
    state prediction, discrete codes, and prior alignment; AgentSociety supplies
    a synthetic operationalization rather than independent psychological
    validation.
\end{itemize}

\section{Related Work}

\paragraph{Trait--state modeling and behavioral simulation.}
Psychological theories distinguish relatively stable between-person traits from
within-person states that fluctuate across time and context. Longitudinal
evidence further shows that trait and state motivation can contribute
independently and interactively to effort-based decisions~\citep{pshyco_background}.
Simulation platforms provide a controlled setting in which such dynamics can be
observed and modeled: Unity-based experimental frameworks reproduce established
behavioral effects~\citep{ufx,Kozlov2010RealBI}, while interview-conditioned
generative agents can reproduce several empirical patterns in human
responses~\citep{park2026llmagentsgroundedselfreports}. Our focus is not to assign a fixed psychological type to an agent, but to use
observed environmental and behavioral state changes as structured supervision
for profile-conditioned behavior prediction.

\paragraph{Personality-conditioned agents.}
Personality-aware agents commonly inject profile information through prompts,
memory, needs, value systems, or policy conditioning. AgentSociety combines
needs, affect, cognition, and memory in large-scale social
simulation~\citep{piao2025agentsociety}, whereas D2A maps personality profiles
to desire-driven values for daily activity generation~\citep{d2a}. Other work
has measured how Big Five traits alter LLM-agent decisions and
performance~\citep{bigfive,duan2025powerpersonalityhumansimulation}, or used
consistency-based optimization to reduce persona drift across
interactions~\citep{abdulhai2025consistentlysimulatinghumanpersonas}. These
methods establish the value of explicit profiles, but primarily use them as
static conditions on behavior. We instead separate the profile from the time-varying environment--behavior
state and model how their interaction informs subsequent behavior.

\paragraph{World models and latent transition learning.}
Latent world models learn compact predictive representations of sequential
environment dynamics and use them for imagination or control. Dreamer introduced
policy learning in a recurrent latent state space~\citep{dreamerv1}, while
discrete and more general successors improved representation learning and
cross-domain robustness~\citep{dreamerv2,dreamerv3}. We adopt the predictive
principle of world modeling for a different scientific target: learning how
questionnaire-derived traits condition environment--behavior state changes. Our
transition-informed posterior uses realized future changes to discover discrete
modes, and a prior predicts those modes from pre-transition context for rollout.
The codes therefore represent transition-level predictive structure rather than
static personality categories.

\section{Method}
\label{sec:method}

\subsection{Problem Formulation: Transition-Informed Behavioral Prediction}
\label{sec:problem}
For participant \(i\), the stable questionnaire-derived profile is
\(\mathbf{q}_i\in\mathbb{R}^{d_q}\). Each behavioral trajectory is represented
by transitions
\[
(\mathbf{s}_t,\mathbf{q}_i,a_t,\mathbf{s}_{t+1},a_{t+1}),
\]
where \(\mathbf{s}_t\in\mathbb{R}^{49}\) is the recorded
environment--behavior state and \(a_t\) is the current macro behavior. Our
primary empirical objective is
\[
p(a_{t+1}\mid\mathbf{q}_i,\mathbf{s}_t,a_t)
\]
for participants whose profiles and trajectories are absent from training.

Rather than predict the next macro behavior only from current context, we
jointly model
\[
p(\mathbf{s}_{t+1},a_{t+1}
\mid\mathbf{q}_i,\mathbf{s}_t,a_t).
\]
The realized next state supplies structured transition supervision during
training. At inference, a predicted state and prior transition code condition
behavior without access to \(\mathbf{s}_{t+1}\). This factorization treats the
latent codes as predictive summaries of profile-conditioned state transitions,
not as causal psychological or clinical constructs.

\begin{figure*}[t]
  \centering
  \includegraphics[width=0.98\textwidth]{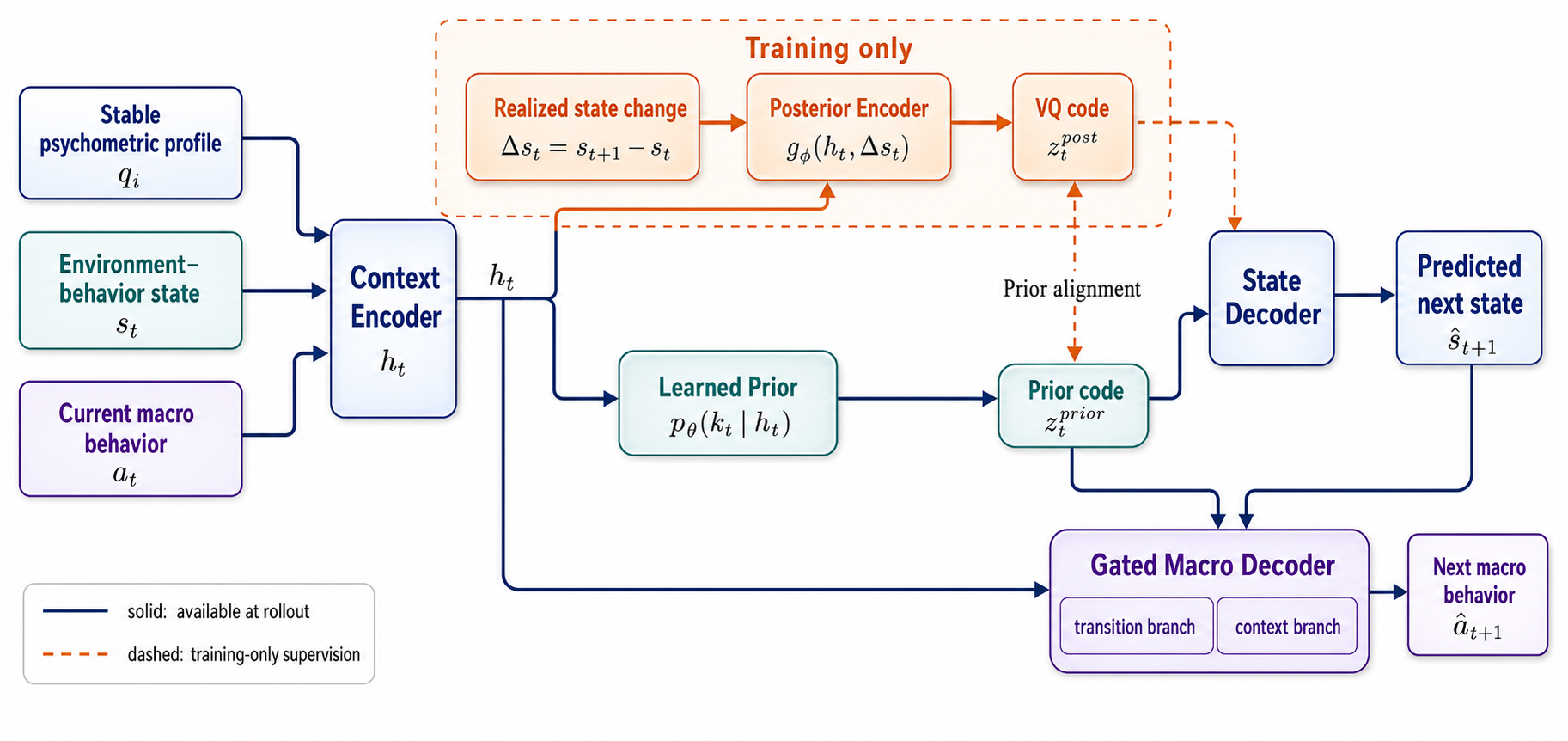}
  \caption{A stable psychometric profile conditions latent transition codes
  together with the current environmental and behavioral state and macro action. The
  predicted next-state representation then informs subsequent behavior prediction.
  Dashed components use the realized future only during training; solid paths are
  available at rollout.}
  \Description{A profile, current environment-behavior state, and current macro
  behavior enter a context encoder. A training-only posterior observes the
  realized state change and supplies a vector-quantized code, while a prior
  predicts a rollout code from pre-transition context. The code conditions a
  next-state decoder and a gated macro-behavior decoder.}
  \label{fig:framework}
\end{figure*}

\subsection{Overview}
The profile-conditioned state world model first encodes the stable profile,
current environment--behavior state, and macro behavior. A training-time
posterior uses the realized future state change to identify a discrete
transition code, while a prior predicts that code from information available
before the transition. Conditioned on the code, the world model predicts the
next state. The transition code and predicted next state feed a macro behavior
decoder, and a simulator-specific execution policy realizes the macro behavior
as primitive actions. The code is therefore a transition-level predictive
variable rather than a participant identity or personality label.

\subsection{Profile-Conditioned State World Model}
\label{sec:worldmodel}
We represent stable and time-varying information separately. The psychometric
profile \(\mathbf{q}_i\) is fixed for participant \(i\), whereas
\(\mathbf{s}_t\) records the time-varying task, environmental, and behavioral
state. The current macro behavior \(a_t\) specifies the behavioral context under
which the transition occurs. For AgentSociety,
\[
\mathcal{A}_{\mathrm{AS}}=
\{\texttt{work},\texttt{consume},\texttt{rest},\texttt{leisure},\texttt{move}\}.
\]

The context encoder fuses these variables:
\[
\mathbf{h}_t =
f_{\theta}^{\mathrm{ctx}}(\mathbf{q}_i,\mathbf{s}_t,a_t).
\]
Given a latent transition code \(\mathbf{z}_t\), the world-model decoder predicts
the next environment--behavior state:
\[
\hat{\mathbf{s}}_{t+1}
=
f_{\theta}^{\mathrm{wm}}(\mathbf{h}_t,\mathbf{z}_t).
\]
This decoder supplies structured transition supervision and makes the
profile--state--behavior factorization explicit instead of using the profile
only as a direct action descriptor.

\subsection{Transition-Informed Latent Representation}
\label{sec:latent_dynamics}
The latent code must be informative about a realized transition yet available
before that transition occurs. A code learned only from current context can
collapse into a static subject embedding or a shortcut for the current state.
A code inferred from the realized future is more transition-specific, but
cannot be used directly at inference.

We address this tension with a prior--posterior scheme. The rollout prior predicts
\[
p_{\theta}(k_t\mid\mathbf{h}_t)
=\mathrm{Cat}\!\left(\mathrm{softmax}(W_z\mathbf{h}_t)\right).
\]
During training and rollout, we use the hard assignment
\(k_t^{\mathrm{prior}}=\arg\max_k p_{\theta}(k\mid\mathbf{h}_t)\) and
\(\mathbf{z}_t^{\mathrm{prior}}=\mathbf{c}_{k_t^{\mathrm{prior}}}\).
During training, the realized state change
\[
\Delta\mathbf{s}_t=\mathbf{s}_{t+1}-\mathbf{s}_t
\]
provides privileged transition supervision. The posterior encoder maps the
pre-transition context and realized change into a continuous latent vector:
\[
\tilde{\mathbf{z}}_t
=
g_{\phi}(\mathbf{h}_t,\Delta\mathbf{s}_t).
\]
The posterior code is available only during training. The hard prior code is also
evaluated during training through prior alignment and next-state reconstruction,
and it is the only code used at rollout.

We quantize the posterior vector with the learned codebook
\(\mathcal{C}=\{\mathbf{c}_j\}_{j=1}^{K}\), where \(K=128\) and
\(\mathbf{c}_j\in\mathbb{R}^{64}\):
\[
k_t^{\mathrm{post}}
=
\arg\min_{j\in\{1,\ldots,K\}}
\|\tilde{\mathbf{z}}_t-\mathbf{c}_j\|_2^2,
\qquad
\mathbf{z}_t^{\mathrm{post}}
=
\mathbf{c}_{k_t^{\mathrm{post}}}.
\]
The VQ objective is
\[
\mathcal{L}_{\mathrm{VQ}}
=
\|\mathrm{sg}[\tilde{\mathbf{z}}_t]-\mathbf{z}_t^{\mathrm{post}}\|_2^2
+
\beta
\|\tilde{\mathbf{z}}_t-\mathrm{sg}[\mathbf{z}_t^{\mathrm{post}}]\|_2^2,
\]
where \(\mathrm{sg}[\cdot]\) is the stop-gradient operator.

The assignment induces the one-hot posterior
\(q_{\phi}(k_t\mid\cdot)=\mathbf{1}\{k_t=k_t^{\mathrm{post}}\}\). The prior
predicts this assignment from \(\mathbf{h}_t\) alone:
\[
\mathcal{L}_{\mathrm{prior}}
=
D_{\mathrm{KL}}\!\left(
q_{\phi}(k_t\mid\cdot)\,\|\,p_{\theta}(k_t\mid\mathbf{h}_t)
\right)
=
-\log p_{\theta}(k_t^{\mathrm{post}}\mid\mathbf{h}_t).
\]
The prior-alignment arrow in Fig.~\ref{fig:framework} denotes this
cross-entropy. The posterior discovers transition-informed codes from observed
futures, while the prior predicts them using only pre-transition information.

\subsection{Transition-Conditioned Macro Behavior Prediction}
\label{sec:macro}
The world model predicts both the next state and the next macro behavior. The
latent code summarizes how the current profile, environment--behavior state,
and macro behavior are associated with the next transition. We compute the
transition-conditioned logits as
\[
\ell_{t+1}^{\mathrm{trans}}
=
f_{\theta}^{\mathrm{trans}}
(\mathbf{h}_t,\mathbf{z}_t,\hat{\mathbf{s}}_{t+1}).
\]
A context branch provides a complementary direct prediction:
\[
\ell_{t+1}^{\mathrm{ctx}}
=
f_{\theta}^{\mathrm{ctx\text{-}macro}}
(\mathbf{h}_t,\mathbf{s}_t).
\]
We combine the branches with a learned gate:
\[
\gamma_t
=
\sigma\!\left(
f_{\theta}^{\mathrm{gate}}
(\mathbf{h}_t,\mathbf{z}_t,\hat{\mathbf{s}}_{t+1})
\right),
\qquad
\ell_{t+1}
=
\gamma_t\ell_{t+1}^{\mathrm{trans}}
+
(1-\gamma_t)\ell_{t+1}^{\mathrm{ctx}}.
\]
This design makes macro prediction transition-aware without requiring the
latent branch when the current context is already sufficient.

\subsection{Joint Training Objective}
We jointly train state prediction, vector quantization, prior alignment, and
macro behavior prediction:
\[
\mathcal{L}
=
\lambda_s\mathcal{L}_{\mathrm{state}}
+
\lambda_{\mathrm{VQ}}\mathcal{L}_{\mathrm{VQ}}
+
\lambda_z\mathcal{L}_{\mathrm{prior}}
+
\lambda_a\mathcal{L}_{\mathrm{macro}}.
\]
During training, the shared decoder forms
\(\hat{\mathbf{s}}_{t+1}^{r}=f_{\theta}^{\mathrm{wm}}(\mathbf{h}_t,\mathbf{z}_t^{r})\)
for \(r\in\{\mathrm{post},\mathrm{prior}\}\). The equally weighted objective is
\[
\mathcal{L}_{\mathrm{state}}
=\sum_{r\in\{\mathrm{post},\mathrm{prior}\}}
\|\hat{\mathbf{s}}_{t+1}^{r}-\mathbf{s}_{t+1}\|_2^2.
\]
The alignment loss trains the prior distribution despite hard selection, while
prior reconstruction exposes the decoder to the rollout code. At rollout, only
the prior-code prediction is used. For the observed
next macro behavior \(a_{t+1}^{\star}\),
\[
\mathcal{L}_{\mathrm{macro}}
=
-\log p_{\theta}
(a_{t+1}^{\star}\mid\mathbf{q}_i,\mathbf{s}_t,a_t).
\]
The gated decoder is trained through \(\mathcal{L}_{\mathrm{macro}}\) applied
to the combined logits. All variants use the same class-balanced macro loss.
The VQ-based variants additionally use a lightweight code-usage regularizer to
mitigate codebook collapse.

\subsection{Simulation-Grounded Execution}
\label{sec:micro}
After a macro behavior is predicted, the execution policy realizes it as a
short-horizon trajectory segment. The macro behavior specifies the high-level
intent, while the micro controller instantiates that intent under the current
simulator state. Let \(\xi_{t:t+H}\) denote the executable segment over horizon
\(H\). The execution policy is
\[
\pi_{\omega}^{\mathrm{exec}}
(\xi_{t:t+H}\mid\mathbf{q}_i,\mathbf{s}_t,a_t).
\]
Here, \(\mathbf{q}_i\) preserves participant-level conditioning,
\(\mathbf{s}_t\) grounds execution in the simulator, and \(a_t\) specifies the
macro intent.

We train the execution policy with DPO~\citep{dpo}. Under the same context,
pairs are ranked by deviation from the corresponding human trajectory:
\(\xi^{+}\) is the closer match and \(\xi^{-}\) has substantially larger
deviation. The objective is
\begin{equation}
\begin{aligned}
\mathcal{L}_{\mathrm{DPO}}
= -\log\sigma\Bigg(\beta\Bigg[
&\log\frac{
\pi_{\omega}^{\mathrm{exec}}(\xi^{+}\mid\mathbf{q}_i,\mathbf{s}_t,a_t)}
{\pi_{\mathrm{ref}}(\xi^{+}\mid\mathbf{q}_i,\mathbf{s}_t,a_t)}
\\[-1mm]
-&\log\frac{
\pi_{\omega}^{\mathrm{exec}}(\xi^{-}\mid\mathbf{q}_i,\mathbf{s}_t,a_t)}
{\pi_{\mathrm{ref}}(\xi^{-}\mid\mathbf{q}_i,\mathbf{s}_t,a_t)}
\Bigg]\Bigg),
\end{aligned}
\end{equation}
where \(\pi_{\mathrm{ref}}\) is the reference execution policy and \(\beta\)
controls preference strength.

\section{Experiments}

Experiments examine whether transition-informed supervision improves behavior
prediction for unseen participants, how continuous next-state modeling,
discretization, and prior alignment contribute, and whether the same
profile--state--behavior factorization can be operationalized in a structurally
different simulator. VEET provides the primary human-participant evaluation;
AgentSociety provides a synthetic cross-simulator test.

\subsection{Experimental Setup}

\paragraph{Simulation Environment.}
We evaluate our framework on two complementary simulation platforms:
\textbf{AgentSociety}~\cite{piao2025agentsociety} and \textbf{VEET}, a self-built
Unity-based behavioral simulator. AgentSociety is a synthetic, LLM-driven city-scale
multi-agent simulation platform in which agents maintain internal memory, update
needs and internal simulator states, generate daily activity plans, and execute behaviors
through modular action blocks. It provides a synthetic testbed for operationalizing
profile-conditioned behavior under a different state and action interface.
VEET is a Unity-based simulator designed to study behavioral trade-offs between
reward acquisition and anxiety- or effort-related pressure. Further environment details appear in Section~\ref{appendix:veet}. Participants make
sequential navigation and task decisions in a controlled interactive
environment.
% We collected questionnaire responses and simulation trajectories
% from approximately 300 human participants, yielding paired subject-level
% profiles and behavioral demonstrations for evaluating personalized long-horizon
% behavior generation.

\paragraph{AgentSociety Simulation.}
We use AgentSociety as a synthetic testbed and adapt it to our hierarchical formulation by injecting
questionnaire-derived psychological profiles into each LLM-controlled agent and
constraining its behavior to five macro actions: \texttt{work},
\texttt{consume}, \texttt{rest}, \texttt{leisure}, and \texttt{move}.
Each agent is simulated over a full daily schedule rather
than through isolated one-step queries. At each decision point, the LLM observes
the agent's current needs, profile-derived traits, location, and recent context,
and selects a macro action that is executed over the corresponding simulation
interval.
This yields temporally grounded macro-action traces that can be
directly compared with our hierarchical policy models. Implementation details of this adaptation appear in Section~\ref{appendix:agentsociety}.

\begin{figure*}[t]
    \centering
    
    % 第一张子图（位于上方）
    \begin{subfigure}{1.0\linewidth}
        \centering
        \includegraphics[width=0.8\linewidth]{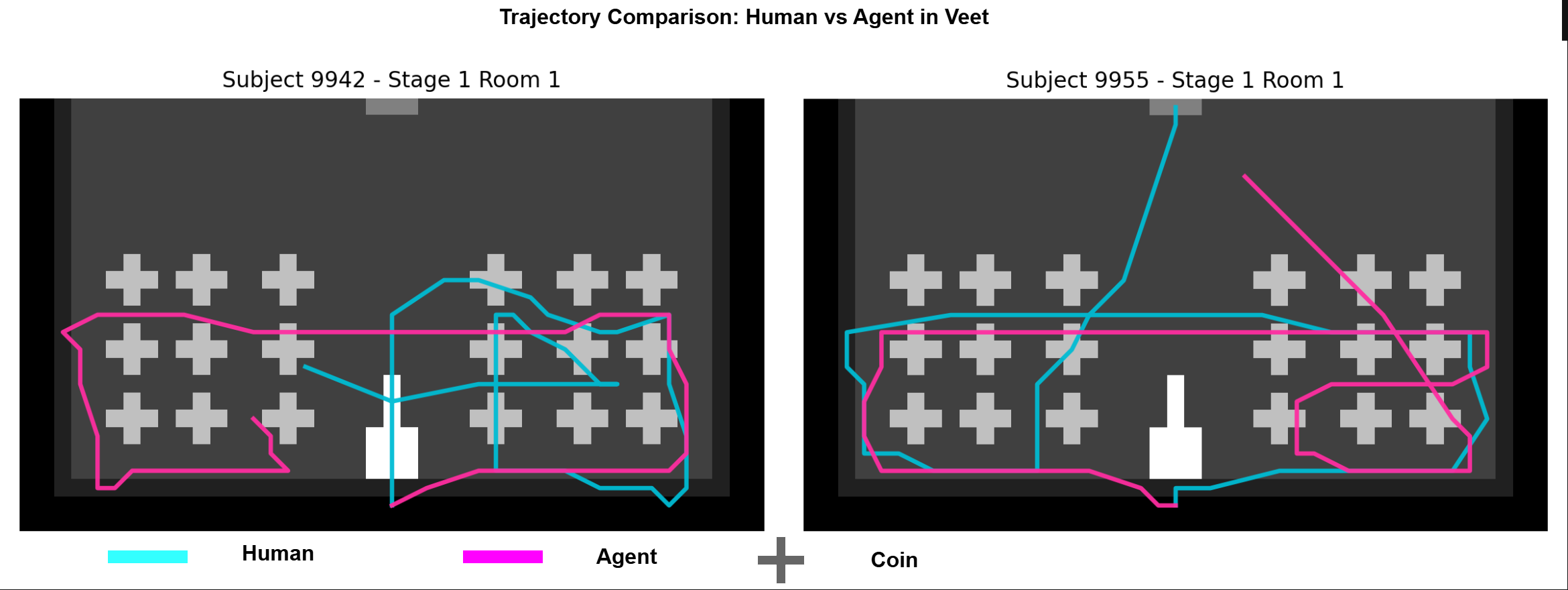}
        \caption{VEET human--agent trajectory comparison} 
        \label{fig:sub_observed}
    \end{subfigure}
    
    \vspace{1em} % 在两张图之间添加一点垂直间距
    
    % 第二张子图（位于下方）
    \begin{subfigure}{1.0\linewidth}
        \centering
        \includegraphics[width=0.8\linewidth]{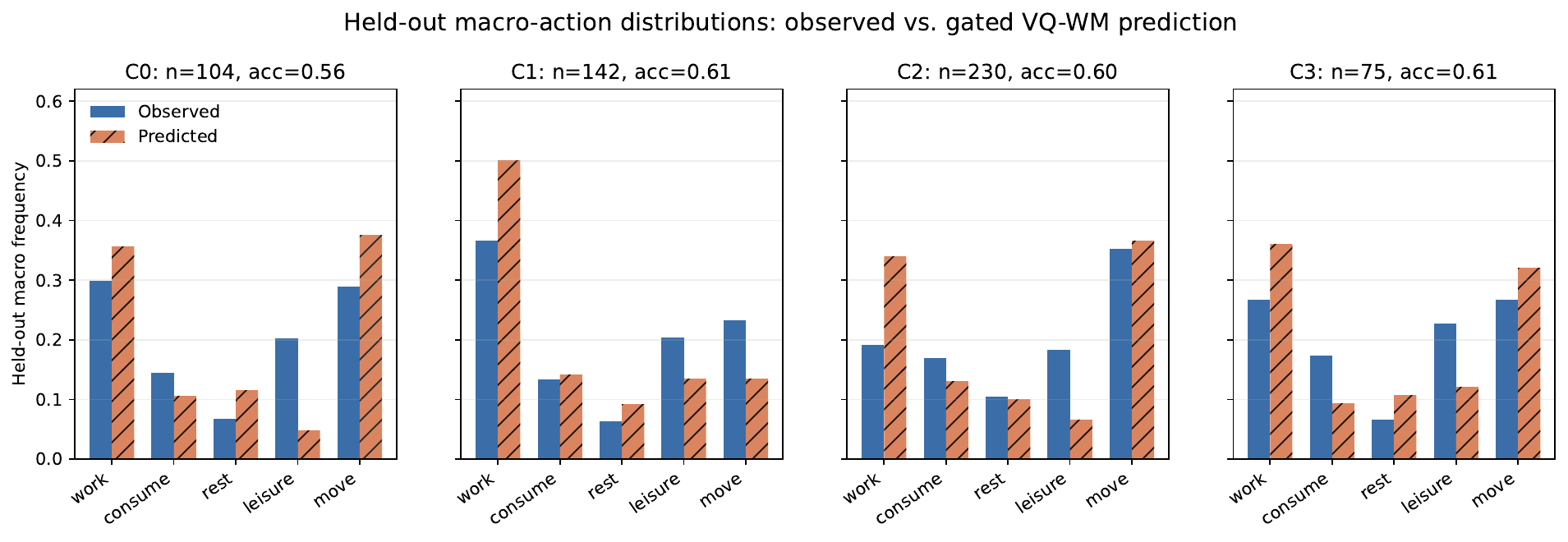}
        \caption{AgentSociety observed vs. predicted macro distributions} 
        \label{fig:sub_true_vs_pred}
        \label{fig:traj}
    \end{subfigure}
    
    \caption{Cross-environment qualitative diagnostics. Panel (a) compares human
    and agent trajectories for two held-out VEET participants. Panel (b) compares
    observed and gated VQ-WM macro-action distributions within AgentSociety-specific
    questionnaire-profile clusters; its compact labels C0--C3 correspond to
    A-C0--A-C3 in Table~\ref{tab:profile_error_groups}. Here, \(n\) denotes the
    number of held-out macro-action events rather than unique participants. The
    panels visualize environment-specific outcomes and are not pooled.}
    \Description{Panel (a) compares VEET human and agent trajectories. Panel (b)
    compares observed and predicted AgentSociety macro-action distributions across
    environment-specific questionnaire-profile clusters.}
\end{figure*}

\begin{figure*}[t]
    \centering
    \includegraphics[width=0.8\textwidth]{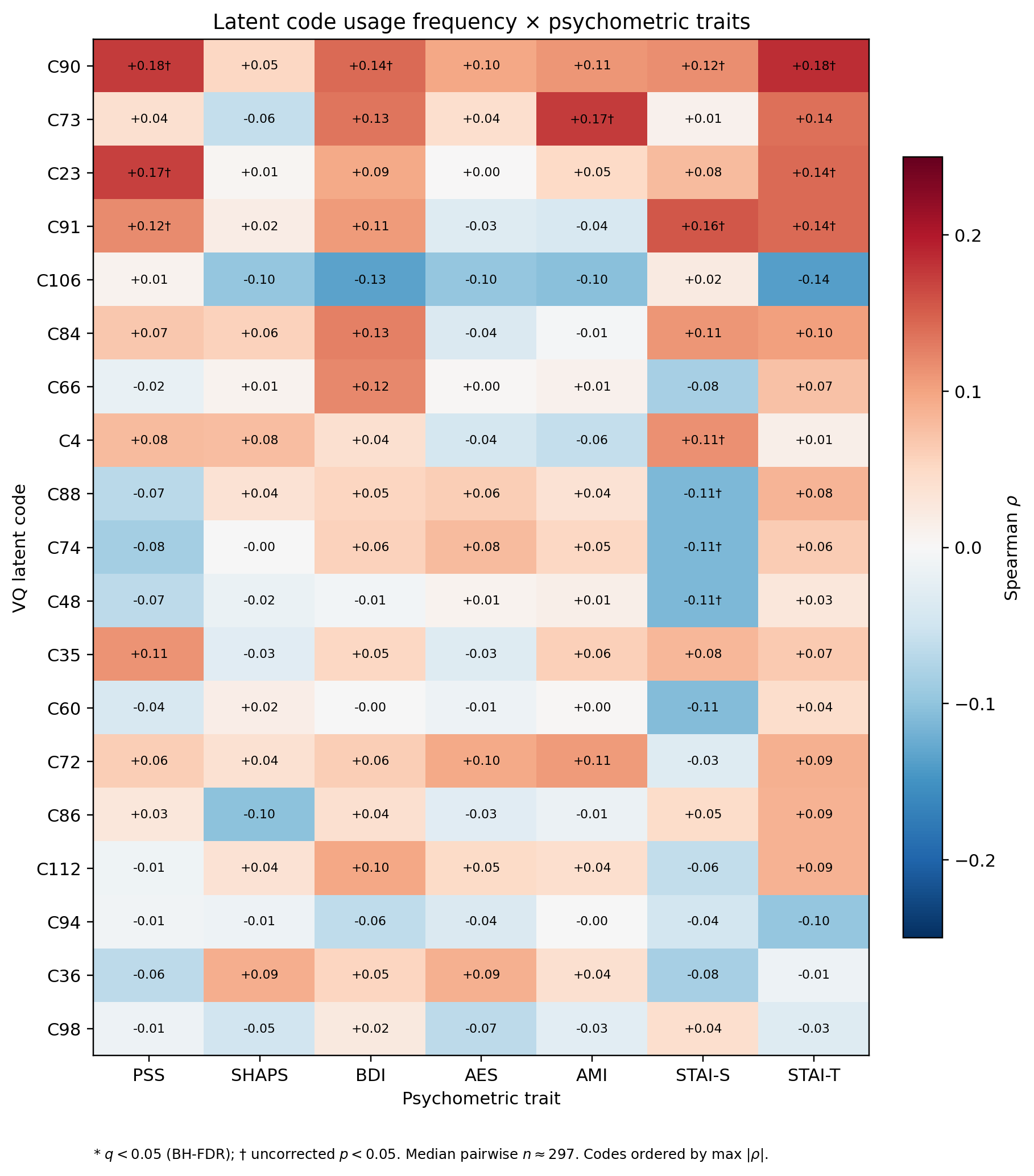}
    \caption{Exploratory associations between participant-level prior-code
    usage frequencies and seven psychometric scales. Cells report Spearman
    $\rho$; $*$ denotes BH--FDR $q<0.05$ across 133 tests and $\dagger$
    denotes uncorrected $p<0.05$. No association survives FDR correction.
    The median pairwise sample size is 297, and codes are ordered by maximum
    absolute correlation.}
    \Description{A diverging heatmap of Spearman correlations between 19 prior
    VQ-code usage frequencies and seven questionnaire scales. Correlations are
    small, and none survives false-discovery-rate correction.}
    \label{fig:code_trait_heatmap}
\end{figure*}
\paragraph{Dataset and Subjects.}
We use questionnaire profiles from 313 VEET participants as the common
subject representation across both environments. In VEET, these participants provide questionnaire profiles and behavioral
trajectories for training the state-transition world model and macro-behavior
predictor. Post-level subjective ratings are used only as auxiliary targets in
the Direct SFT/BC baseline and are not modeled as dynamic affective states by
the proposed world model. For AgentSociety, the same profile
vectors initialize synthetic LLM-controlled agents in the modified simulator. Agents are
rolled out over a one-day schedule starting at 08:00 with a 5-minute base
simulation resolution, and trajectories are logged event-wise when actions are
committed. Each LLM decision is constrained to the structured macro vocabulary
$\{\texttt{work}, \texttt{consume}, \texttt{rest}, \texttt{leisure},
\texttt{move}\}$, yielding profile-conditioned daily behavior traces.

\paragraph{Subject-Level OOD Evaluation.}
We treat the human participant as the domain factor and use a strictly
subject-disjoint split of \textbf{240 training}, \textbf{25 validation}, and
\textbf{48 test participants}. Each participant, questionnaire profile, and
associated trajectory belongs to exactly one split; there is no participant
overlap across training, validation, and test. This protocol therefore tests
whether the behavioral predictor generalizes to unseen people, rather than measuring
performance on new trajectories from previously observed people.
Table~\ref{tab:main_results} reports this subject-level OOD evaluation.

\paragraph{Implementation Details.}
The trajectory encoder \(E_\phi\) is a two-layer transformer with hidden dimension
128 and four attention heads. The VQ codebook contains \(K=128\) entries of
dimension 64, updated via exponential moving average with decay 0.99. The profile encoder \(E_{\psi}^{q}\) is a two-layer MLP with hidden dimension 64 and ReLU activations.
The macro decoder uses a gated two-branch architecture: one branch conditions on
the transition code and predicted next state, and the other on current context; the
logits from the two branches are combined with a learned scalar gate.
All models are trained with Adam
\((\mathrm{lr}=3\times10^{-4}, \beta_1=0.9, \beta_2=0.999)\) for 100 epochs with
batch size 256. The objective weights are set to
\(\lambda_s=1.0\), \(\lambda_{\mathrm{VQ}}=0.25\),
\(\lambda_z=0.1\), and \(\lambda_a=1.0\). All variants use the same class-balanced macro loss. The VQ-based variants
additionally use the code-usage regularizer described in
Section~\ref{sec:method}; the gated branches are optimized through the combined
macro loss. All experiments
are run on a single NVIDIA A100 GPU.

\paragraph{Evaluation Metrics.}
We report Top-1 and Top-3 accuracy for next-macro-behavior prediction. Let
\(P_d\) and \(P_p\) denote the empirical and predicted macro-action
distributions. Macro KL is \(D_{\mathrm{KL}}(P_d\|P_p)\), and
macro-distribution similarity is
\(1-\mathrm{JSD}(P_d,P_p)/\log 2\). These metrics evaluate distributional
alignment beyond the dominant class; the JSD-based measure does not by itself
establish long-range temporal dependence. We also report the number of unique
VQ codes activated by the prior as a diagnostic of codebook collapse.
\begin{table*}[t]
\centering
\small
\caption{
Main results under subject-level OOD evaluation on VEET and AgentSociety.
The 240/25/48 train/validation/test participants are mutually exclusive, so
the test set evaluates generalization to unseen people rather than new
trajectories from people observed during training.
Macro Top-1/Top-3 evaluate next macro-action prediction; Macro KL is
$D_{\mathrm{KL}}(P_{\mathrm{data}}(a)\|P_{\mathrm{pred}}(a))$, and
Dist. Sim. is the macro-action distribution similarity
$1-\mathrm{JSD}(P_{\mathrm{data}},P_{\mathrm{pred}})/\log 2$.
For Ours on VEET, the second line reports 95\% participant-level bootstrap
confidence intervals over all 48 test participants (1,000 resamples; seed 2026).
}
\label{tab:main_results}
\begin{tabular}{lccccc}
\toprule
Model
& Top-1 $\uparrow$
& Top-3 $\uparrow$
& Macro KL $\downarrow$
& Dist. Sim. $\uparrow$
& VQ Codes $\uparrow$ \\
\midrule
\multicolumn{6}{c}{\textbf{VEET}} \\
\midrule
Random
& 0.167
& 0.500
& 0.937
& 0.666
& -- \\
Majority
& 0.739
& --
& 6.888
& 0.855
& -- \\
Direct SFT / BC
& 0.754
& 0.890
& 0.361
& 0.846
& -- \\
IRL policy
& 0.650
& 0.845
& 0.431
& 0.758
& -- \\
% VQ-WM
% & \textbf{0.905}
% & 0.981
% & 0.0078
% & 0.987
% & \textbf{23} \\
Ours
& \shortstack{\textbf{0.904}\\{\scriptsize [0.895, 0.913]}}
& \shortstack{\textbf{0.989}\\{\scriptsize [0.987, 0.990]}}
& \shortstack{\textbf{0.0033}\\{\scriptsize [0.0023, 0.0046]}}
& \shortstack{\textbf{0.999}\\{\scriptsize [0.998, 0.999]}}
& 19 \\
\midrule
\multicolumn{6}{c}{\textbf{AgentSociety}} \\
\midrule
Random
& 0.200
& 0.600
& 0.081
& 0.869
& -- \\
Majority
& 0.298
& --
& 0.843
& 0.504
& -- \\
Direct SFT
& 0.525
& 0.826
& 0.158
& 0.681
& -- \\
% VQ-WM
% & 0.555
% & 0.897
% & 0.184
% & 0.948
% & 2 \\
% VQ-WM + balanced loss
% & 0.568
% & \textbf{0.938}
% & \textbf{0.052}
% & \textbf{0.981}
% & \textbf{7} \\
Ours
& \textbf{0.595}
& 0.933
& 0.078
& 0.974
& 5 \\
\bottomrule
\end{tabular}
\end{table*}
\subsection{Unseen-Participant Behavioral Prediction}
Table~\ref{tab:main_results} evaluates the complete 48-participant test set,
with no participant, profile, or trajectory shared with training. On VEET, the
profile-conditioned world model reaches 0.904 Top-1 and 0.989 Top-3 accuracy,
compared with 0.754 and 0.890 for direct SFT/BC. It also reduces Macro KL to
0.0033 and reaches 0.999 macro-distribution similarity. The
participant-level 95\% bootstrap interval for Top-1 is [0.895, 0.913], with
the remaining intervals reported directly in the table. Together, these
results show that the learned transition representation supports macro-behavior
prediction for previously unseen participants; they do not establish causal
psychological dynamics.

\begin{table}[ht!]
\centering
\small
\caption{
VEET latent representation ablation.
}
\label{tab:ablation_veet_simple_vs_vq}
\begin{tabular}{lccc}
\toprule
Variant
& Top-1 $\uparrow$
& Top-3 $\uparrow$
& Macro KL $\downarrow$ \\
\midrule
Simple WM + decoder
& 0.901
& 0.952
&  0.012\\
VQ-WM
& 0.905
& 0.981
& 0.0078 \\
VQ-WM + prior align.
& 0.904
& 0.989
& \textbf{0.0033} \\
\bottomrule
\end{tabular}
\end{table}
\subsection{Contribution of Transition Supervision}
Table~\ref{tab:ablation_veet_simple_vs_vq} separates continuous next-state
prediction, discrete transition modeling, and rollout-time prior alignment. The
largest pointwise difference is between direct SFT/BC in
Table~\ref{tab:main_results} (0.754 Top-1) and the continuous world model
(0.901), associating explicit next-state supervision with most of the Top-1
gain. Because these variants also differ architecturally, this comparison is
evidence of predictive utility rather than a controlled causal attribution.
Vector quantization changes Top-1 only slightly (0.905) but improves Top-3 from
0.952 to 0.981 and reduces Macro KL from 0.012 to 0.0078. Prior alignment
further lowers Macro KL to 0.0033 and raises Top-3 to 0.989 while preserving
Top-1. Thus, continuous transition supervision supplies the principal
pointwise gain, whereas the discrete prior--posterior design mainly improves
behavioral coverage and distributional alignment while making the
transition-informed representation available before the future is observed.

A profile-cluster diagnostic tests whether this aggregate result is concentrated
in a small questionnaire subgroup. VEET Top-1 remains between 0.891 and 0.913
across the reported groups, although teleport recall is lower and more variable
(0.13--0.26). Because cluster sizes are limited, we treat these differences as
descriptive boundaries rather than subgroup or causal effects
(Table~\ref{tab:veet_profile_error_groups}).

\begin{table}[t]
\centering
\caption{Descriptive held-out VEET performance by environment-specific
questionnaire-profile cluster.}
\label{tab:veet_profile_error_groups}
\scriptsize
\setlength{\tabcolsep}{3pt}
\begin{tabular}{@{}c p{0.42\columnwidth} ccc@{}}
\toprule
Cluster & Profile tag & $N$ & Top-1 & Teleport R. \\
\midrule
V-C0 & Mid profile & 7 & 0.910 & 0.18 \\
V-C1 & Elevated stress & 18 & 0.913 & 0.13 \\
V-C2 & Higher stress / BDI & 8 & 0.905 & 0.26 \\
V-C3 & High hedonic / low BDI & 15 & 0.891 & 0.26 \\
\bottomrule
\end{tabular}
\end{table}

\subsection{What the Latent Modes Represent}
The latent variable is intended to summarize transition-level predictive
structure, not to encode a static personality category. We therefore examine
19 analyzed prior codes against seven psychometric scales (133 tests). No
code--scale association survives BH--FDR correction, and nominal effects are
small ($|\rho|\approx0.11$--$0.19$). This result does not prove
disentanglement, but it is consistent with the transition modes not being
reducible to individual questionnaire dimensions
(Figure~\ref{fig:code_trait_heatmap}).

At the uncorrected level, several small associations provide exploratory structure without supporting psychometric labels. C90 is positively associated with PSS, STAI-T, BDI, and STAI-S; C23 and C91 show related stress--anxiety patterns; C73 is positively associated with AMI; and C88, C74, and C48 are negatively associated with STAI-S. Code-use entropy and the number of unique codes likewise yield no FDR-significant scale association. We therefore treat these patterns as hypothesis-generating diagnostics of transition-mode usage, not as evidence that individual codes correspond to traits, symptoms, or clinical categories.

\subsection{Synthetic Cross-Simulator Operationalization}
AgentSociety tests whether the same profile--state--behavior factorization can
be operationalized in a synthetic LLM-based simulator with a different state
space and action interface. Ours obtains the highest reported Top-1 accuracy (0.595) and
maintains strong macro-distribution alignment in Table~\ref{tab:main_results}.
Figure~\ref{fig:sub_true_vs_pred} further compares observed and predicted
cluster-level action mixtures. The rollouts preserve differences among work,
consumption, leisure, rest, and movement across profile groups, while the
prediction tracks the dominant cluster-level patterns. We interpret this as a synthetic cross-simulator operationalization, not as a
second independent validation of human psychological mechanisms; the AgentSociety agents are initialized from
questionnaire-derived profiles. For descriptive diagnostics, we fit and label
the questionnaire-profile clusters separately within each environment; their
identifiers are therefore environment-specific and are not comparable across
Tables~\ref{tab:veet_profile_error_groups} and~\ref{tab:profile_error_groups}.
AgentSociety implementation details appear in
Appendix~\ref{appendix:agentsociety}.

\begin{table}[t]
\centering
\caption{Descriptive held-out AgentSociety performance by environment-specific
questionnaire-profile cluster.}
\label{tab:profile_error_groups}
\scriptsize
\setlength{\tabcolsep}{3pt}
\begin{tabular}{@{}c p{0.42\columnwidth} ccc@{}}
\toprule
Cluster & Profile tag & $N$ & Top-1 & Leisure R. \\
\midrule
A-C0 & Mid profile & 13 & 0.558 & 0.16 \\
A-C1 & High hedonic / low BDI & 14 & 0.606 & 0.27 \\
A-C2 & Elevated stress & 10 & 0.600 & 0.25 \\
A-C3 & Higher stress / BDI & 11 & 0.613 & 0.24 \\
\bottomrule
\end{tabular}
\end{table}

Table~\ref{tab:profile_error_groups} further tests whether cross-simulator performance is concentrated in one questionnaire-defined subgroup. Top-1 varies from 0.558 to 0.613 across the four clusters, while leisure recall ranges from 0.16 to 0.27. The mid-profile cluster is comparatively harder, whereas the remaining clusters have similar aggregate accuracy despite different stress, BDI, and hedonic profiles. Because these groups contain 10--14 held-out participants, we interpret the differences as descriptive diagnostics rather than causal subgroup effects. The narrow performance range indicates that the aggregate AgentSociety result is not driven by a single profile cluster.

\begin{table}[t]
\centering
\scriptsize
\setlength{\tabcolsep}{3.2pt}
\renewcommand{\arraystretch}{0.92}
\caption{Micro-control results for SFT/BC and macro-conditioned DPO.}
\label{tab:micro_results}
\begin{tabular}{lcccc}
\toprule
& \multicolumn{2}{c}{\textbf{VEET}}
& \multicolumn{2}{c}{\textbf{AgentSoc.}} \\
\cmidrule(lr){2-3}\cmidrule(lr){4-5}
\textbf{Metric}
& \textbf{SFT/BC}
& \textbf{DPO}
& \textbf{SFT/BC}
& \textbf{DPO} \\
\midrule
Completion $\uparrow$
& 0.168
& 0.332
& 0.664
& 0.990 \\
Coin gain $\uparrow$
& 0.120
& 2.516
& --
& -- \\
First-act. KL $\downarrow$
& 2.556
& 1.382
& 1.468
& 0.712 \\
Move. gap $\downarrow$
& 0.873
& 0.520
& 0.453
& 0.267 \\
\bottomrule
\end{tabular}
\end{table}
\noindent Table~\ref{tab:micro_results} defines Completion as the fraction of
rollouts completing the requested macro action, Coin gain as the mean number of
VEET coin rewards per rollout, First-act. KL as the divergence between model and
human first-action distributions, and Move. gap as the absolute difference in
overall \texttt{move} frequency. DPO improves completion and reduces both gaps;
on VEET, it also increases coin gain.

\FloatBarrier

\section{Limitations and Ethical Considerations}
The VEET data were collected by the original study team under institutional
human-subjects oversight at Emory University, with informed consent from all
participants. This work uses only de-identified data and involved no new
participant recruitment or contact; access was restricted to authorized
researchers.

Our evaluation establishes the downstream predictive utility of transition
supervision; it does not independently assess calibrated multi-step
state-transition fidelity. The model does not contain a separately observed
dynamic affect channel; its transition representation is learned from
environmental and behavioral variables recorded by VEET. The learned codes are
behavior-predictive
summaries rather than causal, clinical, or diagnostic states, and the model is
not intended for consequential decisions about individuals. Participant-disjoint
evaluation addresses transfer to unseen people within this study design;
validation across additional populations and instruments remains future work.
Responsible use requires de-identification, restricted access, and human
oversight.

\section{Conclusion}
We study transition-informed supervision for behavior prediction in unseen
participants. The world model separates stable profiles, current states, and
behavior: a posterior extracts codes from observed changes, while a prior
predicts them before the future is available. VEET results associate next-state
modeling with pointwise accuracy and discretization and prior alignment with
behavioral coverage and distributional alignment. AgentSociety provides a
synthetic cross-simulator operationalization. These results support the world
model as a predictive inductive bias rather than a validated psychological
mechanism.

\section{Generative AI Usage}
Generative AI tools assisted manuscript organization and language editing,
LaTeX and figure preparation, and development and debugging of analysis code.

\newpage

\bibliographystyle{ACM-Reference-Format}
\bibliography{custom}

\appendix
\section{AgentSociety}
\label{appendix:agentsociety}
\subsection{Modifications}
We use AgentSociety as a synthetic testbed and modify it to expose a
profile-conditioned hierarchical action interface. The modification consists of four components: profile injection,
need-dynamics modification, block-level action restriction, and trajectory
logging.

\textbf{Profile injection.}
For each subject, we first convert the questionnaire vector into an
\texttt{AgentPsychProfile}. The profile contains initial need satisfactions,
emotion values, personality text, and three composite behavioral dimensions:
hedonic capacity, effort capacity, and stress level. After the simulator creates
the native \texttt{SocietyAgent} instances, we patch each live agent by writing
these values into its memory state. Specifically, we initialize
\texttt{hunger\_satisfaction}, \texttt{energy\_satisfaction},
\texttt{safety\_satisfaction}, \texttt{social\_satisfaction}, and the added
\texttt{pleasure\_satisfaction} field, together with the agent's emotion state,
emotion types, background text, and initial thought.

\textbf{Need-dynamics modification.}
We extend the original need state with a pleasure-related dimension. This
dimension is initialized from questionnaire-derived hedonic traits, especially
measures related to anhedonia and momentary pleasure. We also patch the
\texttt{NeedsBlock} parameters after agent construction. Questionnaire traits
scale the decay rates of hunger, energy, safety, social, and pleasure needs
through multiplicative factors
$\alpha_H,\alpha_D,\alpha_P,\alpha_C,\alpha_L$. Anxiety- and hedonic-related
traits further shift the safety and pleasure thresholds. This makes the same
environment state induce different dominant needs for different profile types.
For example, high stress can make safety needs trigger earlier, while low
hedonic capacity can make pleasure satisfaction decay faster and trigger
leisure-seeking behavior sooner.

\textbf{Social-behavior suppression.}
Because our experiments focus on individual profile-conditioned behavior rather
than interpersonal dynamics, we suppress the original social pathway. We remove
social behavior from the plan guidance options, replace the detailed planning
prompt with a no-social version, and set the social satisfaction state to a high
value while making its decay nearly zero. We also remove any registered
\texttt{SocialBlock} from the agent dispatcher when present. This prevents the
agent from spending trajectory steps on messaging, calling, or meeting other
agents, and ensures that logged behaviors correspond to individual daily
activities.

\textbf{Structured macro-action interface.}
The original AgentSociety planner can generate free-form natural-language
actions. We replace this with a five-category macro-action vocabulary:
$\mathcal{A}=\{\texttt{work},\texttt{consume},\texttt{rest},
\texttt{leisure},\texttt{move}\}$. Internally, these correspond to coarse plan
types \texttt{work}, \texttt{consume}, \texttt{sleep}, \texttt{leisure}, and
\texttt{mobility}. The detailed plan prompt is constrained so that each plan
step must use exactly one of these coarse types and must avoid fine-grained
sub-actions. We additionally introduce or expose dedicated rest and leisure
execution paths so that \texttt{rest} and \texttt{leisure} correspond to
executed behavior modules rather than merely textual descriptions.

The macro-action prior is profile-conditioned. The current dominant need
provides a base preference over macro actions: hunger favors consumption,
tiredness favors rest, safety favors work, and pleasure favors leisure. The
composite profile dimensions then shift these logits: hedonic capacity increases
the tendency toward consumption and leisure, effort capacity increases work and
mobility, and stress increases rest while suppressing consumption, leisure, and
movement. The resulting logits define a profile-conditioned macro distribution
used both for simulation analysis and as the behavioral abstraction aligned with
our hierarchical policy models.

\begin{algorithm}[t]
\caption{AgentSociety profile-conditioned simulation}
\label{alg:agentsociety-profile}
\begin{algorithmic}[1]
\Require Questionnaire profiles $\{q_i\}$, selected subjects $\mathcal{S}$
\For{$i\in\mathcal{S}$}
    \State $p_i \gets \textsc{QuestionnaireToProfile}(q_i)$
    \State Create AgentSociety agent $a_i$
    \State Inject $p_i$ into memory needs, emotions, personality, and thought
    \State Patch \texttt{NeedsBlock} decay rates and thresholds
    \State Remove social guidance and disable \texttt{SocialBlock}
    \State Restrict plan steps to
    $\{\texttt{work},\texttt{consume},\texttt{rest},
    \texttt{leisure},\texttt{move}\}$
\EndFor
\For{each simulation time step $t$}
    \For{each agent $a_i$}
        \State Observe current simulator state, location, memory, and profile
        \State Compute profile-conditioned macro prior
        \State LLM selects a structured macro action and generates a plan
        \State Execute the corresponding AgentSociety behavior block
        \State Log state, profile, macro action, probabilities, and description
    \EndFor
\EndFor
\end{algorithmic}
\end{algorithm}

\textbf{Execution and logging.}
Each LLM-controlled agent is simulated over a daily schedule. At each discrete
simulation time point, the agent observes its current need state, emotion,
thought, profile text, location, and recent memory context. The LLM selects a
guidance option and expands it into a short structured plan whose steps belong
to the allowed macro categories. During post-processing, each executed memory
stream event is mapped back to one of the five macro actions. We log the
simulation tick, current need state, satisfaction values, questionnaire profile,
macro action, macro-action logits and probabilities, and the corresponding
micro-action description. This produces event-level trajectories that are
temporally grounded in the AgentSociety simulator and directly comparable to
the macro-action sequences used by our hierarchical world-model and policy
components.

\paragraph{AgentSociety trajectory generation details.}
For AgentSociety, each rollout uses the VEET questionnaire-derived profile
vector $\mathbf{q}_i$ to initialize the agent's native need and emotion
fields. The profile determines hedonic capacity, effort capacity, perceived
stress, and need-specific decay or threshold parameters. Simulations begin at
08:00 and run over a one-day schedule. AgentSociety advances with
\texttt{ticks\_per\_step}=300, corresponding to a 5-minute simulation
resolution. However, logged trajectories are event-based rather than uniformly
sampled: a record is written when the agent commits an action in its memory
stream. As a result, logged behavioral intervals are variable, typically ranging
from 5 to 30 simulated minutes.

At each decision point, the LLM observes the agent's current native simulator
state, location, profile-derived traits, and recent memory context. It first
selects a profile-conditioned guidance option and then expands it into a
structured plan of up to six coarse steps. Each step is constrained to one of
five macro-action categories:
$\{\texttt{work}, \texttt{consume}, \texttt{rest}, \texttt{leisure},
\texttt{move}\}$. These event-level macro-action traces form the AgentSociety
dataset used for profile-conditioned behavior modeling.
\section{VEET Explanation}
\label{appendix:veet}

\subsection{VEET Environment and Training State Details}
\label{app:veet_details}

\paragraph{VEET environment.}
VEET is an embodied Unity-based behavioral task in which participants navigate
through a multi-room virtual environment, collect coins, move between rooms
through doors, and occasionally use teleportation. The environment records both
low-level motor behavior and task-level progress variables. Each participant
also provides questionnaire responses measuring affective and motivational
traits. These questionnaire responses define a subject profile vector
$\mathbf{q}_i$, which conditions our models throughout training.

\paragraph{Observation state.}
At each time step, we convert the raw Unity observation into a flat numerical
state vector $\mathbf{s}_t$. The state contains four groups of information:
(1) self-motion and position, including player position, rotation, movement
distance, and wall distances; (2) room context, including current room, room
effort, room reward, door positions, and distances to front and back doors;
(3) object layout, including remaining coin counts and nearest-coin relative
positions/distances; and (4) task progress, including collected coins, points,
button presses, teleport counts, elapsed time, room time, round time, and
timing flags for post-level subjective queries, but not the subjective rating
values. The raw state vector is normalized with the
same feature scaler used by the behavioral cloning and reward-model baselines.
When clean-feature checkpoints are used, we apply the corresponding feature
mask before normalization so that all models use an identical state interface.

\paragraph{Questionnaire conditioning.}
Each subject $i$ is associated with a questionnaire-derived vector
$\mathbf{q}_i$. Depending on the experiment, $\mathbf{q}_i$ is represented either
as item-level responses, compressed scale-level features, or eight scale scores
corresponding to PSS, SHAPS, BDI, AES, AMI, MAPSR, STAI-S, and STAI-T. The same
$\mathbf{q}_i$ is concatenated with the environment state or encoded as profile
context in the policy and world-model networks. Thus, the model predicts
behavior not only from the current physical state, but also from stable
subject-level psychometric traits.

\paragraph{Macro-action labels.}
We derive a six-way macro-action label from human replay trajectories.

These labels are inferred from geometric progress, movement speed, button
presses, room effort, and short-horizon events. For example,
\texttt{go\_to\_coin} corresponds to movement that reduces distance to nearby
coins, \texttt{go\_to\_next\_door} and \texttt{go\_to\_prev\_door} correspond
to room-transition-directed behavior, and \texttt{teleport} corresponds to
explicit teleport usage. This macro abstraction provides the high-level action
space for our hierarchical policy and world model.

\paragraph{Micro-action space.}
The lower-level VEET controller uses the native discrete Unity action space.
Each action is represented as a triplet
$(m_t, r_t, z_t)$, where $m_t$ is movement intensity, $r_t$ is turn direction,
and $z_t$ indicates teleportation. Non-teleport actions enumerate movement and
turn combinations, while a separate discrete action is reserved for teleport.
This gives a 13-way micro-action space. The macro-conditioned micro policy
receives the normalized state $\mathbf{s}_t$, the subject profile
$\mathbf{q}_i$, and an embedding of the current macro action, then outputs a
distribution over these 13 low-level actions.

\paragraph{Training transitions.}
We distinguish the state-world-model and micro-control interfaces. The former
uses
\[
\mathcal{D}_{\mathrm{WM}}=
\{(\mathbf{s}_t,\mathbf{q}_i,a_t^{\mathrm{macro}},\mathbf{s}_{t+1},
a_{t+1}^{\mathrm{macro}})\},
\]
whereas the micro controller uses
\[
\mathcal{D}_{\mathrm{micro}}=
\{(\mathbf{s}_t,\mathbf{q}_i,a_t^{\mathrm{macro}},
a_t^{\mathrm{micro}},\mathbf{s}_{t+1})\}.
\]
This separation prevents the observed micro action from entering the world-model
transition tuple. For macro-conditioned DPO, candidate trajectory segments are
initialized from the same replay context and conditioned on the macro label. We
form preference pairs by comparing candidates with the corresponding human replay
trajectory using progress signals such as coin gain, door progress, movement
distance, room change, and teleport events. The closer candidate is preferred,
whereas a segment with substantially larger behavioral deviation is rejected.

\paragraph{Role in our framework.}
VEET therefore provides both human demonstration trajectories and
questionnaire-conditioned subject profiles. The trajectories supervise
macro-action prediction, micro-action execution, and latent transition
modeling, while the profiles allow the model to capture individual differences
in strategy selection. This makes VEET a controlled testbed for evaluating whether transition-informed
supervision supports profile-conditioned behavior prediction beyond a
majority-action shortcut.

\end{document}